\documentclass[pdflatex,sn-mathphys-num]{sn-jnl}% Math and Physical Sciences Numbered Reference Style
\usepackage{graphicx}%
\usepackage{multirow}%
\usepackage{amsmath,amssymb,amsfonts}%
\usepackage{amsthm}%
\usepackage{mathrsfs}%
\usepackage[title]{appendix}%
\usepackage{xcolor}%
\usepackage{textcomp}%
\usepackage{manyfoot}%
\usepackage{booktabs}%
\usepackage{algorithm}%
\usepackage{algorithmicx}%
\usepackage{algpseudocode}%
\usepackage{listings}%
\theoremstyle{thmstyleone}%
\theoremstyle{thmstyletwo}%

\theoremstyle{thmstylethree}%

\begin{document}

\title[Classification and detection of multiple UAVs using rational Gaussian wavelet neural networks]{Classification and detection of multiple UAVs using rational Gaussian wavelet neural networks}


\author[1,2]{\fnm{Gerg\H{o}} \sur{Ungv\'ari}}\email{aele02@inf.elte.hu}

\author[6]{\fnm{Ferenc} \sur{Braun}}\email{braun.ferenc@ek.hun-ren.hu}
\author[1,5]{\fnm{Attila} \sur{\'Amon}}\email{ze3vjn@inf.elte.hu}
\author[3]{\fnm{P\'eter} \sur{Kackst\"adter}}\email{kackstadter.peter@broadcast.hu}
\author[6]{\fnm{J\'anos} \sur{Volk}}\email{volk.janos@ek.hun-ren.hu}
\author*[1]{\fnm{P\'eter} \sur{Kov\'acs}}\email{kovika@inf.elte.hu}
\author[2,4]{\fnm{Tam\'as} \sur{D\'ozsa}}\email{tamas.dozsa@unidistance.ch}


\affil*[1]{\orgdiv{Department of Numerical Analysis}, \orgname{E\"otv\"os Lor\'and University}, \orgaddress{\street{Pázmány Péter sétány}, \city{Budapest}, \postcode{1117}, \country{Hungary}}}

\affil[2]{\orgdiv{System and Control Laboratory}, \orgname{HUN-REN Institute for Computer Science and Control}, \orgaddress{\street{Kende utca}, \city{Budapest}, \postcode{1111}, \country{Hungary}}}

\affil[3]{\orgname{Szent György Hang- és Filmművészeti Technikum}, \orgaddress{\street{Lenhossék utca}, \city{Budapest}, \postcode{1096}, \country{Hungary}}}

\affil[4]{\orgdiv{Department of Mathematics and Computer Science}, \orgname{UniDistanceSuisse}, \orgaddress{\street{Schinerstrasse}, \city{Brig-Glis}, \postcode{3900}, \country{Switzerland}}}

\affil[5]{\orgname{Siemens Mobility Kft.}, \orgaddress{\street{Gábor Dénes utca}, \city{Budapest}, \postcode{1117}, \country{Hungary}}}

\affil[6]{\orgname{HUN-REN Centre for Energy Research}, \orgaddress{\street{Konkoly-Thege Miklós út}, \city{Budapest}, \postcode{1121}, \country{Hungary}}}

\abstract{The detection of unmanned aerial vehicles (UAVs) is important for the protection of civilian and military infrastructure. In this paper we propose a cost effective UAV detection system using sound signals obtained from microphones. The recorded signals are passed through a signal processing pipeline which employs interpretable adaptive feature extractors using so-called rational Gaussian wavelets. These adaptive wavelet transformations are embedded into and trained together with an underlying small neural network which detects and classifies UAVs based on the obtained features. This leads to a physically interpretable machine learning algorithm that in addition to classifying UAVs is also capable of detecting UAV swarms. We demonstrate our results using data collected in indoor studio and noisy outdoor environments. We conclude that the proposed method outperforms traditional machine learning approaches for detecting and classifying single UAVs as well as drone swarms, while retaining a high degree of interpretability.  Our implementation of the proposed methods is made publicly available for reproducibility. }




\maketitle

\section{Introduction}\label{sec1}

The detection and classification of unmanned aerial vehicles (UAVs) has become an important problem in recent years. UAVs, also commonly known as drones, have been successfully applied for the delivery of goods, agriculture and various other civilian industries~\cite{seidaliyeva2023advances}. On the other hand, exploiting their ability to carry weaponry and surveillance equipment, UAVs have found a large array of military applications and play an ever increasing role on the modern battlefield~\cite{DronesBattle,swinney2022review,wang2021counter}. In addition, drones provide a cost efficient method for criminal groups to carry out illegal activities such as smuggling~\cite{samaras2019deep, DedroneWorldwideDroneIncidents} and disrupting industry and air traffic~\cite{seidaliyeva2023advances, DedroneWorldwideDroneIncidents}.  We recommend~\cite{seidaliyeva2023advances} and~\cite{droneRev2} for a deeper discussion on the different threats posed by malicious UAV activity. 

In order to counter the threat posed by drones, the development of trustworthy and affordable technology to detect and classify UAVs is necessary. Accordingly, several recent works consider the problem of drone detection using a variety of technologies. The detection and classification of drones is a difficult measurement and signal processing task for a number of reasons. Depending on the type of the UAV, its sound, radar or visual signatures can vary widely~\cite{mohsan2023unmanned}. In addition, the measurement apparatus usually needs to operate in an outside environment subject to sudden changes in weather conditions and significant background noise~\cite{mohsan2023unmanned}. The appearance of so-called drone swarms, where multiple UAVs perform a coordinated task, further complicates potential defense strategies~\cite{seidaliyeva2023advances}. 

Current state-of-the-art technologies~\cite{seidaliyeva2023advances, droneRev2} consider signals obtained from different types of sensors accompanied by a diverse array of signal processing methods to detect drone activity. Due to the above highlighted difficulties, it is generally accepted that a high performance UAV detection system should incorporate a variety of signals obtained from different sensors. The design of such a system can also depend on other factors, such as the type of infrastructure it is intended to protect, the overall cost of the system and specifics related to the environment (such as predictable weather conditions, noise pollution levels, etc.). Nevertheless, drone detection subsystems that depend on a single mode of measurement are an important field of study, as these provide the components of an overall drone detection system.

Current UAV detection and classification technologies can be divided into four classes depending on the type of measurements they rely on. Radar based methods (see e.g.,~\cite{radar,tanveer2025from}) use radio waves to detect objects. Advantages of this technology include robustness against weather conditions, long range and being able to detect the speed and direction of UAVs~\cite{seidaliyeva2023advances}. Unfortunately, the detection of smaller drones is a difficult task for radar systems due to small radar cross sections and low flying altitudes. In addition, radar based systems are costly due to the complexity of the involved instrumentation.

Vision based detection systems usually employ image processing to analyze recordings from cameras~\cite{camera}. These are very cost effective and can provide visual confirmation of UAVs, however their performance can be hindered by visibility and weather constraints. In addition, a visual detection system is only able to detect UAVs in its line of sight. 

Another interesting family of detection methods attempts to recognize drone activity by capturing and analyzing wireless signals. Methods belonging to this family include~\cite{RF1, RF2}. Benefits of radio frequency analysis include a long operating range and the ability to classify different types of UAVs. On the other hand these methods cannot identify autonomous drones~\cite{seidaliyeva2023advances} or drones controlled by non-wireless means such as carbon-optic fibers~\cite{carbon}.

Finally, we mention UAV detection systems that operate by recoding sound in the surrounding area~\cite{sound1, sound2}. These systems are easy to deploy and are cost effective. Additional advantages of sound based detection schemes include the ability to classify recognized drones, estimate the position of UAVs~\cite{seidaliyeva2023advances} and the ability to operate without a line of sight to the target. Difficulties associated with acoustic drone detection systems include background noise filtering and vulnerability to wind conditions.

In this study, we consider a novel sound based drone detection system. We show that through the use of appropriate signal processing methods, the proposed system is able to
\begin{enumerate}
    \item Mitigate the effect of wind and other background noise, by learning frequency signatures characteristic of UAVs of interest,
    \item Classify UAV types based on the learned signatures,
    \item Recognize drone swarms, that is, distinguish between intrusions by a single UAV and groups of drones.
\end{enumerate}
In this study, we assume that the drones-to-be-detected, are small, electrical vehicles that are traditionally difficult to recognize~\cite{seidaliyeva2023advances, droneRev2}. To reduce the cost of the system, we employ microphones based on micro-electro-mechanical systems (MEMS) technology and a mathematically justified, so-called model driven machine learning (ML) paradigm. 

The use of the proposed novel signal processing scheme clearly distinguishes our approach from previous sound based UAV detection technologies. Although previous drone detectors employing artificial intelligence methods on sound signals have been introduced in~\cite{sound1} and~\cite{sound2}, these rely on traditional machine learning architectures. While deep learning based approaches achieve high performance~\cite{soundDeep}, model parameters do not carry physical meaning. Thus, the output of such models is not explainable. This poses concerns for safety-critical applications such as drone detection. In addition, the size of deep neural models often requires specialized hardware for real-time use, which can lead to increased deployment and maintenance costs.

In contrast, certain previous methods~\cite{sound1} apply static feature extraction steps to obtain meaningful information from recorded sound signals. The extracted features are then passed to different ML models, which are usually smaller than their deep learning counterparts~\cite{soundDeep}. Features are often extracted using time-frequency transformations, and thus the obtained information is interpretable. On the other hand, previous methods only considered static feature extractors. That is, these methods apply a fixed transformation (e.g., short time Fourier transform~\cite{sound1}) to each sound segment. This leads to sub-optimal signal representation, because the feature extractor methods cannot adapt to different UAV types, maneuvers and changing environmental conditions. 

In contrast, our proposed approach relies on adaptive feature extraction transformations embedded into small neural networks. The proposed model is suitable for real time applications on limited hardware and extracts physically meaningful information from the sound signals. Furthermore it provides feature transformations that can adapt to different sound signatures and changing environmental factors. 

The proposed signal processing pipeline falls into the category of so-called model driven machine learning approaches. These mathematically justified ML models are fully, or partially interpretable and can be used in safety-critical applications. Importantly, such methods retain the generalization abilities of classical deep learning approaches, at least in the context of the application to which they are deployed. A number of model driven ML methods have been recently introduced, including variable projection based neural networks~\cite{kovacs2022VPNet} and kernel methods~\cite{vpsvm}. Other recent examples of model driven ML methods include wavelet convolution neural networks~\cite{wang2021automatic, li2021waveletkernelnet}.

In the current study, we propose a new wavelet convolution layer (and a corresponding model driven neural network) that uses so-called rational Gaussian wavelet (RGW) kernels~\cite{RGW}. RGW is a recently introduced class of admissible wavelets, whose morphology can be greatly influenced through a number of parameters. The proposed layer can learn the shape of the optimal mother wavelet, as well as a finite number of scales (corresponding to pseudo frequencies~\cite{daubechies1992ten}) that can be used to detect and classify UAVs. 

Below, we summarize the most important novelties of our study
\begin{enumerate}
    \item We propose a novel, RGW convolution based model driven neural network. The proposed signal processing model obtains physically meaningful wavelet coefficients from the recorded sound signals. These adaptively extracted features are then processed by a small neural network to detect UAV presence and classify UAV types.
    \item We demonstrate that our proposed wavelet kernel convolution based machine learning model can be used to detect the presence of groups of UAVs.
    \item We compare the proposed signal processing approach to several baseline machine learning methods. We show that in addition to the inherent interpretability of the proposed model, it significantly outperforms traditional ML methods for drone detection. All of our experiments are completely reproducible using our publicly available python implementation of the proposed methods (see data and code statement at the end of the article).
    \item We propose a cost-effective, sound based system for UAV detection and classification with a light weight, interpretable ML based signal processing unit.
    %\item We provide a new collection of indoor and outdoor measurements comprising a total of 152 minutes of sound recordings from a number of microphones. The recordings include sound signatures from six different drone types as well as significant background noise (such as helicopters in the case of our outdoor measurements).
\end{enumerate}

The rest of this paper is organized as follows. In section~\ref{sec:daq}, we described the hardware used to collect our dataset and the different measurement scenarios. Section~\ref{sec:sic} provides an overview of the most important properties of the collected sound signals. These properties provide justification for the proposed signal processing pipeline. In section~\ref{sec:rgw}, we review some important properties of rational Gaussian wavelets~\cite{RGW} and introduce interpretable RGW convolution layers. These are then used to construct a small, partially interpretable neural network for drone detection. Section~\ref{sec:exp} details our experiments and results. Finally, in section~\ref{sec:conc} we summarize our findings and discuss future research directions. 

\section{Data acquisition}
\label{sec:daq}

In the studio environment, an Audio/Video Recorder (Video Devices PIX 270i) received the HD-SDI signal from a wide-angle camera and simultaneously generated a 48 kHz word clock signal that was phase-locked to the video reference. Audio recordings were captured at a sampling rate of 192 kHz with a 24-bit resolution in signed Linear Pulse Code Modulation (LPCM) format. The word clock from the PIX 270i was distributed to a Master Clock Generator (Apogee Big Ben), which operated as a frequency multiplier to derive a 192 kHz reference. The Big Ben provided this synchronized 192 kHz word clock to an Audio Interface System (Focusrite RedNet) and the associated Digital Audio Workstation (Pro Tools), ensuring a common time base across all microphone channels. The PIX 270i also recorded the video feed along with an analog reference from one of the microphones, allowing precise post-synchronization between the audio and video domains. The microphones used in the studio were Brüel \& Kjær 4006 and Schoeps MK2 with CMC5 preamplifiers, all of which are pressure-type electroacoustic transducers, ensuring the most phase-coherent data acquisition possible.

To simulate real environmental conditions relevant to the intended system application, a microphone array equipped with commercial MEMS transducers was deployed outdoors. The recordings were carried out at a sampling rate of 8 kHz with a 16-bit resolution in signed LPCM format. The acoustic environment included intense background noise sources such as helicopter overflights, nearby road traffic, human speech, and strong wind. As the four microphones were mounted on a common PCB and shared identical signal-conditioning and acquisition circuitry, their outputs were recorded in temporal synchronization relative to each other.

The experiments were conducted using several commercially available multirotor UAVs, representing different size classes, acoustic signatures, and propulsion configurations. The DJI Mavic Pro and Mavic Pro 2 are foldable quadcopters with a takeoff weight of approximately 734 g and 907 g, respectively. Both platforms use four two-blade rotors with typical hover-speed ranges between 5000–7800 RPM.

The DJI Mavic Mini is an ultralight (249 g) quadcopter equipped with four two-blade rotors, operating at higher fundamental rotor frequencies due to its smaller propeller diameter and reduced inertia.

For outdoor measurements, additional UAV types were included, such as the DJI Mavic 3 Pro (958 g, 4 rotors), the DJI Avata 2 (cinewhoop-style ducted quadcopter, ~377 g, high-RPM 3-blade rotors), and the DJI Matrice 30T, a larger industrial platform weighing approximately 3.7 kg and equipped with four large-diameter two-blade rotors rotating at significantly lower RPM ranges (2500–4200 RPM).

These platforms cover a broad acoustic spectrum: lightweight high-RPM systems produce higher-frequency harmonic structures, while heavier drones exhibit stronger low-frequency components. This diversity enables robust evaluation of the proposed signal processing and classification methodology.
\begin{table}[h]
\centering
\begin{tabular}{lcccc}
\hline
\textbf{UAV Model} & \textbf{Mass} & \textbf{Rotors} & \textbf{Blades/rotor} & \textbf{Typical RPM} \\
\hline
DJI Mavic Pro & 734 g & 4 & 2 & 5000--7800 \\
DJI Mavic Pro 2 & 907 g & 4 & 2 & 5000--7800 \\
DJI Mavic Mini & 249 g & 4 & 2 & 6000--8500 \\
DJI Mavic 3 Pro & 958 g & 4 & 2 & 4800--7200 \\
DJI Avata 2 & 377 g & 4 (ducted) & 3 & 9000--12000 \\
DJI Matrice 30T & 3.7 kg & 4 & 2 & 2500--4200 \\
\hline
\end{tabular}

\vspace{2mm}
\caption{Technical specifications of the UAV platforms used in the study}
\label{tab:uav_specs}
\end{table}

\section{Signal description}
\label{sec:sic}
%% TODO: UG
%% KEY POINTS:
% - Recorded sound data description (sampling rate, channel number, did we use all of them?, etc.)
% - Some examples of frequency change when different types of drones appear and/or swarm appears (spectrograms/scalograms -> this gives some theoretical reasoning for using the wavelet transform).

\begin{figure}[h]
    \centering
    \includegraphics[width=0.99\linewidth]{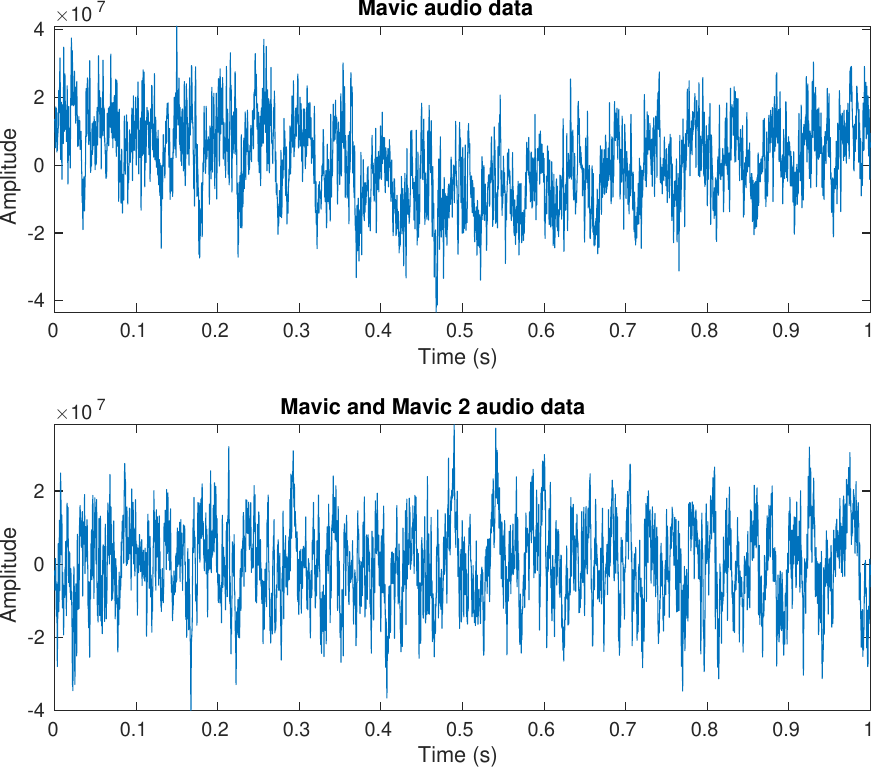}
    \caption{Audio signals of a Mavic drone (left) and a Mavic and Mavic 2 drones flying at 1 meter high.}
    \label{fig:tds}
\end{figure}

We review some important properties of the sound signals obtained from the front-facing Schoeps MK2 microphones described in section~\ref{sec:daq}. Audio recorded from a single such microphone was used to generate the results presented in Section~\ref{sec:exp} relating to the indoor measurements, thus the input to the proposed methods are 1D signals (see Fig.~\ref{fig:tds}). We note that even though the sound signals obtained from our noisy outdoor measurement scenario are worse quality, the characteristics discussed here remain true for them, which is also reflected by our experimental results (see section~\ref{sec:exp}). The signals considered here have been recorded with a sampling rate of 192 kHz. The audio data is split into $100$ ms long, non-overlapping segments. Thus input samples consisted of arrays containing 19200 amplitude values each. The only other preprocessing step applied to the recorded sound signals was normalization, so that the values in every segment had a mean of $0$ and standard deviation $1$.

\begin{figure}
    \centering
    \includegraphics[width=\linewidth]{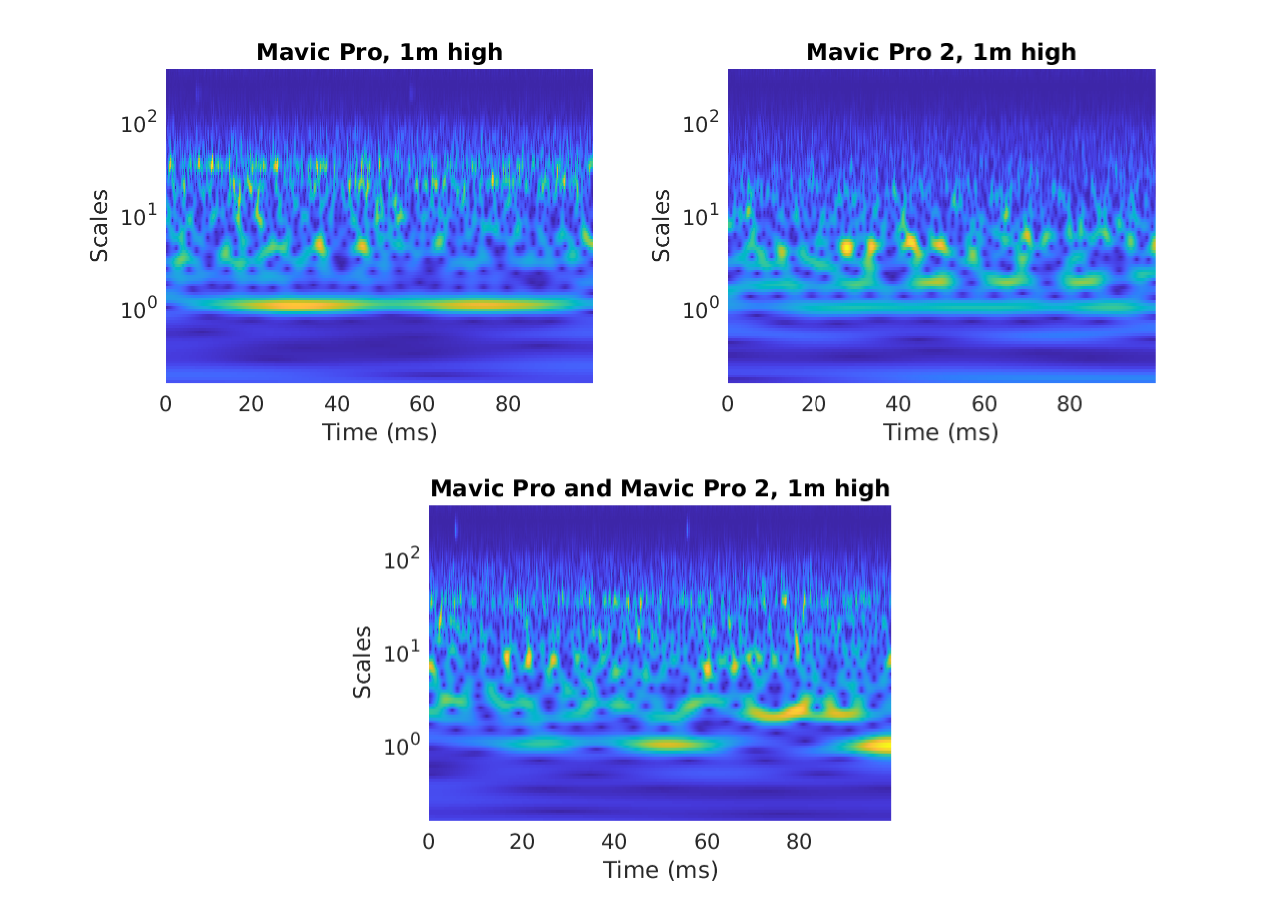}
    \caption{Time-scale representation of audio signals of the Mavic Pro and Mavic Pro 2 drones taken indoors, with very little noise present. The top left signal is the Mavic Pro flying at 1 meter high and the top right is the Mavic Pro flying at 4 meters. The bottom is both the Mavic and Mavic 2 flying at 1 meter. The time-scale representation have been obtained via continuous wavelet transform, using Morle-wavelets.}
    \label{fig:cwts}
\end{figure}

As noted in~\cite{seidaliyeva2023advances, sound1, sound2, soundDeep}, the frequency profiles of the audio signals carry noticeable differences, characteristic of each drone model. This can also be seen in Fig.~\ref{fig:cwts}, as certain features appear consistently in the spectrogram, or the time-frequency representation obtained from the Morlet-transform of signals, only when audio from a Mavic Pro type drone is present.

The characteristics of these features, however, can be subject to change depending on factors other than the model, such as the position, or the movement of the drone. Furthermore, the measured signal segments do not exhibit periodic behavior. If the transformations used during the classification pipeline are not translation invariant, the position (in time) of the identifying features may be subject to change as well.

These qualities of the audio signals mean that drone detection is a nonlinear classification problem of nonstationary and nonperiodic signals. In section~\ref{sec:rgw} we introduce an ML model based on adaptive rational Gaussian wavelet  convolution operators, that can successfully process these types of signals. The experimental results of section~\ref{sec:exp} demonstrate that the scheme provides an apt solution to the identification and classification problem set in this paper.

\section{Rational Gaussian Wavelet convolution networks}
\label{sec:rgw}
%% TODO: UG/AA
%% KEY POINTS:
% - review of wavelet transforms and RGW

Given a signal $f$ and function $\psi$, the continuous wavelet transform of the signal is defined as

\begin{equation}
    W_{\psi}(\lambda,\tau) = \int_{-\infty}^{+\infty} f(t)\overline{\psi}_{\lambda,\tau}(t)dt,
    \label{eq:cwt}
\end{equation}
where
\begin{equation*}
  \psi_{\lambda,\tau}(t) := \frac{1}{\sqrt{\lambda}}\psi  \left(\frac{1}{\lambda}(t-\tau) \right) \quad (t,\lambda,\tau \in \mathbb{R},\lambda \neq 0).
\end{equation*}

We refer to the function $\psi$ as the mother wavelet, while $\lambda$ and $\tau$ denote so-called dilation and translation parameters, respectively. Given fixed parameters $\lambda$ and $\tau$, the number $W_{\psi}(\lambda,\tau)$, is called a wavelet coefficient and it describes the similarity between $f$ and $\psi_{\lambda, \tau}$. We shall assume $\psi, f \in L_2(\mathbb{R})$ henceforth, which is a sufficient condition for~\eqref{eq:cwt} to exist. Furthermore, if the mother wavelet $\psi$ satisfies the so-called admissibility property~\cite{daubechies1992ten}, then the transformation~\eqref{eq:cwt} is invertible (in the $L_2(\mathbb{R})$-sense). The $W_{\psi}$ coefficients describe the so-called time-scale representation of $f$ which is closely related to time-frequency representations~\cite{daubechies1992ten}.

In ~\cite{RGW} the authors introduce a family of wavelet functions, called rational Gaussian wavelets (RGW). RGWs are defined with a parameter vector $\boldsymbol{\eta} := \left[ t_1,t_2,...,t_p,z_1,z_2,...,z_n \right]$ as
\begin{equation}
    \psi^{\boldsymbol{\eta}}(t) = C(\boldsymbol{\eta}) P^{\boldsymbol{\eta}}(t) v^{\boldsymbol{\eta}}(t) e^{\frac{-t^2}{2}} \quad (t \in \mathbb{R}, \boldsymbol{\eta} \in \mathbb{C}^{p+n}),
    \label{eq:rgw}
\end{equation}
where $C$ is a constant that depends only on $\boldsymbol{\eta}$.
%\begin{equation*}
%    C(\boldsymbol{\eta}) = \frac{1}{||\boldsymbol{\eta}||_2}.
%\end{equation*}
The rational term $P^{\boldsymbol{\eta}} (t) v^{\boldsymbol{\eta}}(t)$ is defined as 
\begin{equation}
\label{eq:polterm}  P^{\boldsymbol{\eta}}(t) = t \prod^{p}_{k=1} (t-t_k)(t+t_k), \quad (t_k \in \mathbb{C} \backslash \{0\}, p \in \mathbb{N})
\end{equation}
and $v^{\boldsymbol{\eta}} \in \mathcal{V}$, where
\begin{multline}
\label{eq:rterm}
    \mathcal{V} := \\ \left\{ v(t) = \frac{1}{\prod^{n-1}_{k=0}(t-z_k)(t+z_k)(t-\tilde{z}_k)(t+\tilde{z}_k)}, n \in \mathbb{N} \right\}
\end{multline}
and
\begin{equation*}
    \tilde{z} := -\Re(z)+i\Im(z).
\end{equation*}

\begin{figure}
    \centering
    \includegraphics[width=0.99\linewidth]{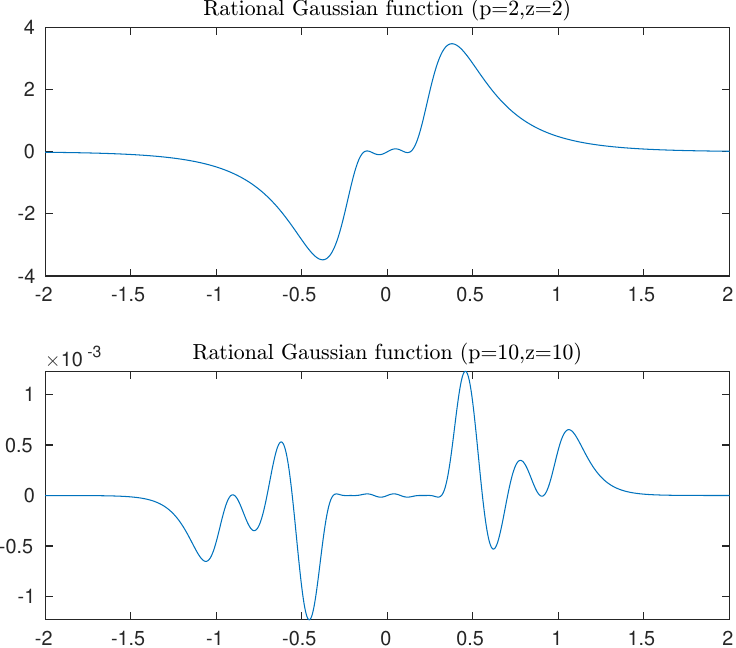}
    \caption{Rational Gaussian functions of Eq.~\eqref{eq:rgw}, with parameters $p$ and $z$.}
    \label{fig:rfs}
\end{figure}

For simplicity, in this work we assume $\{t_k\}_{k=1}^{n} \subset \mathbb{R}$. In~\cite{RGW} the admissibility of $\psi^{\boldsymbol{\eta}}$ is proven, thus (in theory) $f$ can be reconstructed from its RGW wavelet coefficients. Fig.~\ref{fig:rfs} illustrates some RGW mother wavelets with different $\boldsymbol{\eta}$ parameter choices. Fig.~\ref{fig:rfs} also demonstrates how the large number of parameters offers a high degree of flexibility for influencing the morphology of the mother wavelet. This property of RGW wavelets is also well exploited in section~\ref{sec:exp} of this study.

% - Use of RGW as convolution kernels -> novelty

In~\cite{li2021waveletkernelnet} a wavelet based convolution kernel has been introduced, using a fixed mother wavelet, with the translation and dilation pairs as learnable parameters. The approach in~\cite{li2021waveletkernelnet} differs from the one introduced in this article in two regards. First, our use of RGWs allows us to optimize the morphology of the mother wavelet. Secondly, our proposed RGW convolution layer only considers the scales $\lambda_k \ (k=1,\ldots,N, \ N \in \mathbb{N})$ as free parameters. The reason for this is that learning exact translation parameters would increase the size of the convolution layer without affecting the model's accuracy in any meaningful capacity.

An important novelty of the current study is the construction of so-called RGW convolution operators. Define the function $\psi_{\lambda}$ as
\begin{equation}
\label{eq:dilwav}
    \psi_{\lambda} (t) := \frac{1}{\sqrt{\lambda}} \psi\left(\frac{t}{\lambda}\right) \quad (t \in \mathbb{R}, \ \lambda > 0).
\end{equation}
The wavelet coefficients $W_\psi (\lambda,\tau)$ from Eq.~\eqref{eq:cwt}, where $(\tau \in \mathbb{R})$ can then be written as
\begin{equation*}
    W_\psi (\lambda,\tau) = (f \ast \psi_{\lambda})(\tau) = \int_{-\infty}^{+\infty} f(t) \overline{\psi_{\lambda}}(t-\tau) dt.
\end{equation*}
In practical cases the signals that have to be processed are usually only available in a discretely sampled form. Let 
$$f_k = f(t_k) \quad  (t_k \in \mathbb{R_+}, \ k = 1,...,N),
$$  and consider the notation $\boldsymbol{f} := (f_1,f_2,...f_N) \in \mathbb{R}^N$. Let furthermore
\begin{equation*}
    \boldsymbol{\psi}^{\boldsymbol{\eta}}_{\lambda, k} := \psi^{\boldsymbol{\eta}}_{\lambda}(t_k) \quad (k = 1,...,M)
\end{equation*}
denote the discrete sampling of the wavelet $\psi_{\lambda}$ (see Eq.~\eqref{eq:dilwav}).

Then, the wavelet coefficients of $f$ can be approximated with the discrete convolution
\begin{equation}
\left[\boldsymbol{f} \ast \boldsymbol{\psi}_{\lambda}^{\boldsymbol{\eta}}\right]_k := \sum_{j=1}^k \boldsymbol{f}_j \boldsymbol{\psi}^{\boldsymbol{\eta}}_{\lambda, k-j+1}.
    \label{eq:dconv}
\end{equation}

Previous applications of RGW wavelets considered approximating wavelet coefficients using variable projection operators~\cite{golub1973differentiation, golub2003separable}. This approach however heavily depends on the periodic, or quasi-periodic property of the signal. As shown in Fig.~\ref{fig:tds} and in Fig.~\ref{fig:cwts}, the audio signals considered in this paper are not periodic, thus variable projection based computation of RGW coefficients will not capture meaningful features. To overcome this issue, we propose discrete convolution layers with RGW kernels, that are
\begin{enumerate}
    \item capable of extracting meaningful features from non-periodic signals using discrete convolutions,
    \item allow for a high degree of adaptivity of the mother wavelet's morphology due to the nature of RGW.
\end{enumerate}

The key component of the proposed model driven neural network is the RGW convolution layer. This layer applies discrete convlutions (see Eq.~\eqref{eq:dconv}) to the $1D$ input signals, where the kernel $\boldsymbol{\psi}^{\boldsymbol{\eta}}_{\lambda}$ is an RGW wavelet characterized by a dilation parameter $\lambda > 0$ and the parameters in $\boldsymbol{\eta} \in \mathbb{R}^{p+n}$. This vector contains the zeros $t_k \ (k=1,\ldots,p)$ of the polynomial term from Eq.~\eqref{eq:polterm} and the poles $z_j \ (j=1,\ldots,n)$ from Eq.~\eqref{eq:rterm}. Together they greatly influence the shape of the RGW mother wavelet and can be collected to a single vector by
\begin{equation*}
    \begin{split}
        & \boldsymbol{\eta} = (t_1,t_2,...,t_p,z_1,z_2,...,z_n) \in \mathbb{C}^{n+p}\\
        & \quad (n,p \in \mathbb{N}).
    \end{split}
\end{equation*}
The transformation characterizing the proposed RGW layer is defined by
\begin{equation}
    \boldsymbol{f} \rightarrow T_{\boldsymbol{\eta}}\boldsymbol{f} = 
    \begin{bmatrix}
        \boldsymbol{f} \ast \boldsymbol{\psi}_{\lambda_1}^{\boldsymbol{\eta}} \\
        \boldsymbol{f} \ast \boldsymbol{\psi}_{\lambda_2}^{\boldsymbol{\eta}} \\
        \vdots \\
        \boldsymbol{f} \ast \boldsymbol{\psi}_{\lambda_m}^{\boldsymbol{\eta}} \\
    \end{bmatrix} =: \begin{bmatrix}
        W_{\psi^{\eta}}(\lambda_1) \\
        W_{\psi^{\eta}}(\lambda_2) \\
        \vdots \\
        W_{\psi^{\eta}}(\lambda_m) \\
    \end{bmatrix} \in \mathbb{R}^{m \times (N+M-1)},
    \label{eq:wcontr}
\end{equation}
where $\boldsymbol{f} \ast \boldsymbol{\psi}_{\lambda_k}^{\boldsymbol{\eta}} \ (k=1,\ldots,m)$ is the discrete convolution defined in~\eqref{eq:dconv}. The Jacobian matrix of~\eqref{eq:wcontr}, with respect to $\boldsymbol{\eta}$, can be easily computed:

\begin{equation}
    \left(\frac{\partial T_{\boldsymbol{\eta}}\boldsymbol{f}}{\partial\boldsymbol{\eta}}\right)_{\lambda_{k}} = \boldsymbol{f} \ast \frac{\partial\psi^{\boldsymbol{\eta}}_{\lambda_k}}{\partial\boldsymbol{\eta}}. \quad (k=1,\cdots, m).
\end{equation}

In ~\cite{RGW}, the authors offer a formula for computing partial derivatives of $\psi^{\boldsymbol{\eta}}_{\lambda_k}$ with regards to $t_k$ and $z_k$. We rely on this formulation to implement the proposed RGW convolution layer.

The convolution layer is followed by a pooling layer, and a fully connected layer. The pooling layer uses a scheme in which, for every $\lambda_k \ (k=1,\ldots,m)$ scale, only the wavelet coefficient of the maximal value is retained.

\begin{equation}
    MP(T_{\boldsymbol{\eta}}f) =     \begin{bmatrix}
        \operatorname{topQ}(|W_{\psi^{\eta}}(\lambda_1)|) \\
         \operatorname{topQ}(|W_{\psi^{\eta}}(\lambda_2)|) \\
        \vdots \\
         \operatorname{topQ}(|W_{\psi^{\eta}}(\lambda_m)|) \\
    \end{bmatrix} \in \mathbb{R}^{m \times Q},
    \label{eq:maxpool}
\end{equation}
where the operator $\operatorname{topQ}$ selects the $Q \in \mathbb{N}$ largest elements of the argument vector. This allows the proposed neural network architecture to retain physically meaningful information in the following sense. Notice that the $k$-th component of $MP(T_{\boldsymbol{\eta}}f)$ encodes the $Q$ maximum similarity scores achieved between $\boldsymbol{f}$ and $\boldsymbol{\psi}_{\lambda_k}^{\boldsymbol{\eta}} \ (k=1,\ldots,m)$. Since for a fixed $\lambda_k$ dilation the vector of wavelet coefficients $W_{\psi^{\eta}}(\lambda_k)$ (see Eq.~\eqref{eq:wcontr}) corresponds to a pseudo frequency, the output of $MP(T_{\boldsymbol{\eta}}f)$ can be interpreted as the maximal amplitude present in the input signal $\boldsymbol{f}$ at the frequency band defined by $\lambda_k$.

\section{Experiments}
\label{sec:exp}

We conducted a number of experiments to demonstrate the effectiveness of RGW convolution networks for audio-based UAV detection. As described in section~\ref{sec:daq}, our dataset includes indoor and outdoor measurement scenarios along with several different types of drones and drone swarms. Accordingly, different experimental scenarios were considered for the detection and classification of a single, or multiple UAVs.
The considered scenarios are described as follows.
\begin{enumerate}
%    \item Detecting the presence of a single UAV in a studio environment. Here the task of the model is to differentiate between audio segments with only very little background noise, and segments where a single UAV can be heard. The UAV models used for this experiment were DJI Mavic Pro, DJI Mavic Pro 2, and DJI Mavic Mini.
    \item Differentiating between the presence of a single UAV, and multiple UAVs. The recordings used for this experiment were taken in a studio environment. The three categories are no drones being present, the presence of a single UAV, and the presence of multiple UAVs (up to three). The UAV models used for this experiment were DJI Mavic Pro, DJI Mavic Pro 2, and DJI Mavic Mini. 
    \item The classification of a single UAV in a studio environment. In this case, the model receives an audio segment with a single UAV present and has to differentiate between three possible models (Mavic Pro, Mavic Pro 2 and Mavic Mini).
    \item Detecting the presence of a single UAV outdoors, in a noisy environment. The categories are the same as in Scenario 1. The UAV models used for this scenario are Avata 2, Matrice 30T, Mavic Mini and Mavic Pro 3.
\end{enumerate}

\begin{table}
    \centering
    \resizebox{1\textwidth}{!}{
    \begin{tabular}{|c|c|c|c|c|c|}
        \hline\textbf{Scenario id.} & \textbf{Task} & \textbf{Location} & \textbf{No. of UAV Types} & \textbf{No. of UAVs} & \textbf{No. of Classes} \\ \hline
        %1. & Detection & Indoor & 3 & 1 & 2 \\ \hline
        1. & Multiple UAV Detection & Indoor & 3 & 3 & 3 \\ \hline
        2. & Classification & Indoor & 3 & 1 & 3 \\ \hline
        3. & Detection & Outdoor & 4 & 1 & 2 \\ \hline
    \end{tabular}
    }
    \caption{Considered experimental scenarios. The number of UAVs refers to the maximum number of individual UAVs present in a single segment, while the number of UAV types refers to the total number used in that scenario.}
\label{tab:experiments}
\end{table}

Table~\ref{tab:experiments} gives a summary of the considered scenarios. The outdoor experiment (as described in section~\ref{sec:daq}) can be regarded as extremely noisy, while the indoor experiments as noise free.

\begin{table}
    \centering
    \resizebox{1\textwidth}{!}{
    \begin{tabular}{|c|c|c|c|c|}
        \hline\textbf{Scenario id.} & \textbf{Location} & \textbf{No. of Categories} & \textbf{Segments Per Category} & \textbf{Segment Length} \\ \hline
        %1. & Indoor & 2 & 900 & 100 ms \\ \hline
        1. & Indoor &  3 & 1790 & 100 ms \\ \hline
        2. & Indoor &  3 & 1000 & 100 ms\\ \hline
        3. & Outdoor &  2 & 1050 & 100 ms\\ \hline
    \end{tabular}
    }
    \caption{Structure of the datasets used in the individual experiments.}
    \label{tab:dset}
\end{table}

In each experiment, the recorded sound samples were subjected to an identical preprocessing regime described in section~\ref{sec:sic}. The proposed ML models were very similar for each measurement scenario as well. The most significant difference, between the models used for indoors and outdoors scenarios, is the size of the used wavelet convolution kernel~\eqref{eq:wcontr}. In case the size of the kernel does not equal $N$, the size of the convolution output will not be $m \times N$, however it is dependant the size of the kernel and the input, and is different in each scenario. The considered neural network is composed of an RGW convolution layer, followed by two fully connected layers with ReLU activation functions before each. The output of the RGW layer is first normalized:

\begin{equation*}
    \frac{W_{\psi^{\eta}}(\lambda_k) - \mathbf{E}(W_{\psi^{\eta}}(\lambda_k))}{\mathbf{Var}(W_{\psi^{\eta}}(\lambda_k))},
\end{equation*}

where $\mathbf{E}$ is the mean and $\mathbf{Var}$ is the variance, then downsampled with a pooling layer. The pooling layer is unique, in that it samples the dominant coefficients for each $\lambda_k \ (k=1,\ldots,m)$ dilation parameter, essentially functioning as a 1-dimensional filter, as described in Eq.~\eqref{eq:maxpool}. The activation function after the final layer is either sigmoid, in the binary classification case, or softmax, in the case of multiple categories.

The number of $p$ and $n$ learnable parameters used to determine the morphology of the RGW mother wavelet was chosen as $p=1$ and $n=10$. In these experiments, we considered $10$ wavelet filters (the number of scales $m=10$).

The classifier block used after the RGW layer consists of a single fully connected layer with $200$ neurons. The model is trained for $300$ epochs with a batch size of $64$.

Each individual experiment was evaluated via 5-fold cross-validation. The training consisted of randomly splitting the total training data 80-to-20 into a training set, which has been used to fit the model parameters and a test set, on which the accuracy of the model was evaluated. This process has been repeated a total of five times for the experiment, and in the following tables the mean, minimum and maximum accuracy can be seen.

Model performance has been compared to several baseline ML methods. For this purpose, we considered the Random Forest (RF) classifier~\cite{pal2005random}, linear and radial Basis kernel SVM (SVM-L and SVM-RBF) classifiers~\cite{chapelle2007training}, as well as the naive Bayes classifier~\cite{NB} (NB). In addition, we conducted experiments on these scenarios using fully connected (FCNN) as well as convolutional (CNN) neural networks. In each scenario the structure of the FCNN model, and the fully connected layers in the CNN model were identical to the structure of the fully connected layers of the RGW-kernel~\eqref{eq:wcontr} model used in that scenario. Furhtermore, the number of channels, and the size of the convolution layer in the used CNN model had been identical to the one used for the RGW convolution layer. The results are shown in the following tables~\ref{tab:results2},~\ref{tab:results3} and~\ref{tab:results4}.

The accuracy of the trained models are calculated by

\begin{equation}
    Acc = \frac{N_{tp}+N_{tn}}{N_{total}},
\end{equation}
where $N_{tp}$ and $N_{tn}$ are the number of true positive and true negative prediction respectively, and $N_{total}$ is the total number of segments in the test set.

In the following subsections we describe each experimental scenario in detail. We note that all of the hyperparameters were determined using a large grid search of the parameter space. To ensure full reproducibility of our results, we direct the reader to the data and code availability statement at the end of the article.

\begin{figure*}
    \centering
    \includegraphics[trim={0 20cm 0 1cm},clip,width=0.99\linewidth]{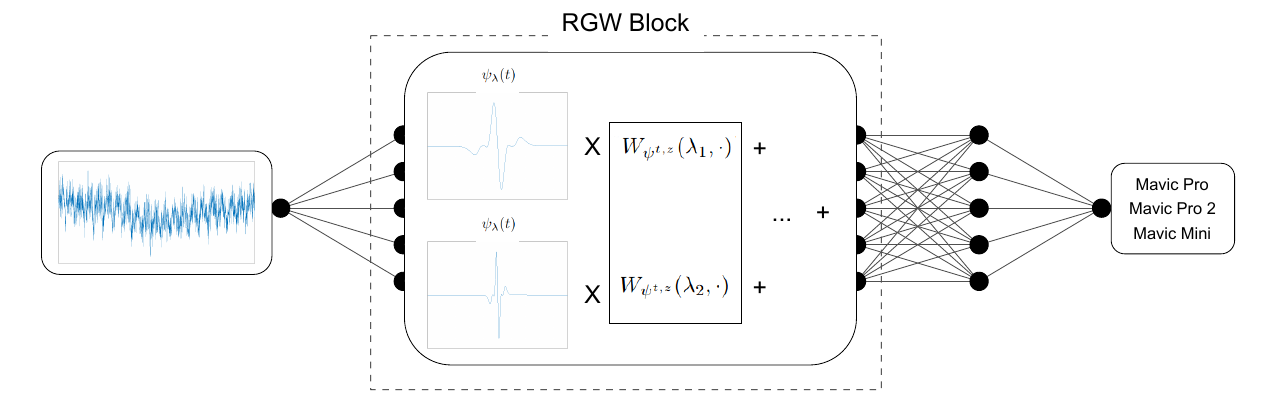}
    \caption{Schematic model of the neural network used for our experiments. The samples are fed first into an RGW convolution layer for feature extraction, then to a fully connected block, used for classification.}
    \label{fig:nn-model}
\end{figure*}

\iffalse
\subsection{Detecting drones in an indoor studio environment}
\label{sec:dronedetect-st}

We proceed to describe experimental scenario 1. from Table~\ref{tab:experiments}. In this scenario, the data split into two categories: one which contains samples with minimal background noise and the other containing samples with sounds from one or more drones. Therefore, the task to detect the presence of at least a single drone can be regarded as a binary classification problem.

\begin{table}[!h]
    \centering
    \begin{tabular}{|c|c|c|c|}
        \hline
        \textbf{Model} & \textbf{Mean Acc.} & \textbf{Min Acc.} & \textbf{Max Acc.}  \\ \hline
        WK-NN (proposed) & $\boldsymbol{97.99 \%}$ & $\boldsymbol{94.69 \%}$ & $ \boldsymbol{99.72} \%$ \\ \hline
        RF & $  93.46\%$ & $ 91.34\%$ & $ 95.53\%$ \\ \hline
        SVM-L & $ 62.12\%$& $  59.50\%$ & $ 65.92\%$ \\ \hline
        SVM-RBF & $ 93.97\%$& $  92.74\%$ & $ 95.25\%$ \\ \hline
        NB & $ 53.46\%$& $  50.27\%$ & $ 55.59\%$ \\ \hline
        CNN & $ 98.88\%$& $  97.77\%$ & $ 99.44\%$ \\ \hline
        FCNN & $ 65.25\%$& $  61.73\%$ & $ 69.83\%$ \\ \hline
    \end{tabular}
    \caption{Results of detecting the presence of a drone in a studio environment (scenario 1. in Table~\ref{tab:experiments}).}
    \label{tab:results1}
\end{table}

 The results are shown in Table~\ref{tab:results1}. As can be seen, apart from the convolutional neural network, the Rational Gaussian Wavelet Kernel-based Neural Network (WK-NN) significantly outperforms the other methods.
\fi

\subsection{Detecting swarms in a studio environment}
\label{sec:swarmdetect-st}

The summary of our experimental scenarios can be found in~\ref{tab:experiments}. In this section we consider the indoors experimental scenario wich we label with id. 1. Our dataset contains three labels: a category containing samples with only background noise, a category with sounds from a single drone, and a category which contains samples with sound recordings from multiple drones. That is, the classification task is to differentiate between a lone drone, background noise, and the presence of multiple drones.

In this particular scenario, the length of each wavelet kernel was determined to be $32$, which results in the size of the convolution output being $10\times 19169$.

The performance of the model has been compared to baseline ML algorithms, as introduced in the beginning of this section. The results can be seen in Table~\ref{tab:results2}. The classical methods, as well as the fully connected neural network, fail to achieve high level of accuracy, while the RGW-kernel based network and the convolutional neural network is able to accurately identify the UAV types. Furthermore, the RGW-kernel based approach overperforms the convolutional neural network. It is important to note that the RGW-kernel based model achieves this accuracy using a lot fewer parameters and the RGW-kernel layer gives us an output whose parameters carry physical meaning about the input signals.

\begin{table}[!h]
    \centering
    \begin{tabular}{|c|c|c|c|}
        \hline
        \textbf{Model} & \textbf{Mean Acc.} & \textbf{Min Acc.} & \textbf{Max Acc.}  \\ \hline
        WK-NN (proposed) & $ \boldsymbol{92.52\%}$& $  \boldsymbol{90.50\%}$ & $ \boldsymbol{96.00\%}$ \\ \hline
        RF & $78.94  \%$ & $ 77.47\%$ & $80.07 \%$ \\ \hline
        SVM-L & $ 50.32\%$& $ 48.42 \%$ & $ 51.96\%$ \\ \hline
        SVM-RBF & $ 90.62\%$& $ 89.20 \%$ & $ 91.62\%$ \\ \hline
        NB & $ 41.49 \%$& $ 39.85 \%$ & $42.64 \%$ \\ \hline
        CNN & $90.57 \%$ & $88.27 \%$ & $92.26\%$  \\ \hline
        FCNN & $ 48.13\%$& $  46.66\%$ & $ 49.03\%$ \\ \hline
    \end{tabular}
    \caption{Results of detecting the presence of a swarm in a studio environment (scenario 1. in Table~\ref{tab:experiments}).}
    \label{tab:results2}
\end{table}

\subsection{Classifying drones in a studio environment}
\label{sec:droneclass-st}

Next, we consider scenario 2. from Table~\ref{tab:experiments}. In the case of identifying drone model types, three categories describe our dataset. Each category contains samples from a single drone, where the categories themselves are characterized by the type of the UAV that can be heard in the sample.

The parameters of the trained model had been the same as the model used in section~\ref{sec:swarmdetect-st}, with the exception of the activation function at the end being Softmax instead of Sigmoid.

The results can be seen in Table~\ref{tab:results3}. Similarly to the previous scenario, the CNN and the RGW-kernel based model are able to achieve a much higher accuracy than the other approaches.

\begin{table}[!h]
    \centering
    \begin{tabular}{|c|c|c|c|}
        \hline
        \textbf{Model} & \textbf{Mean Acc.} & \textbf{Min Acc.} & \textbf{Max Acc.}  \\ \hline
        WK-NN (proposed) & $99.20 \%$ & $98.33 \%$ & $99.83 \%$ \\ \hline
        RF & $  91.53\%$ & $ 90.33\%$ & $ 92.33\%$ \\ \hline
        SVM-L & $ 38.00\%$& $ 36.83\%$ & $ 38.83\%$ \\ \hline
        SVM-RBF & $ 96.37\%$& $  95.17\%$ & $ 97.67\%$ \\ \hline
        NB & $ 48.10\%$& $  46.33\%$ & $ 63.92\%$ \\ \hline
        CNN & $ \boldsymbol{99.97\%}$& $  \boldsymbol{99.83\%}$ & $ \boldsymbol{100.0\%}$ \\ \hline
        FCNN & $ 47.63\%$& $  43.33\%$ & $ 50.83\%$ \\ \hline
    \end{tabular}
    \caption{Results of classifying the UAV model types present in a studio recording (scenario 2. in Table~\ref{tab:experiments}).}
    \label{tab:results3}
\end{table}

\subsection{Detecting drones in a noisy environment}
\label{sec:dronedetect-n}

Finally we consider experimental scenario 3. from Table~\ref{tab:experiments}. For this experiment, our objective is to detect the presence of a single UAV, however the used dataset has been collected in a much less controlled environment, being subject to a large number of noise factors. The details of the noisy dataset are described in section~\ref{sec:daq}.

Just like in the previous examples, the number of dilation parameters learned had been $10$. The numbers $p$ and $n$ of RGW mother wavelet parameters remained $p=1$ and $n=10$, however the size of the convolution kernel was raised to $64$, and the size of the convolution output changed to $10 \times 4347$. As the task is a binary classification problem, Sigmoid had been used as an activation layer.

Table~\ref{tab:results4} contains the accuracy of the RGW-based model, as well as the accuracy of baseline ML methods, also used in the previous experiments. As can be seen, the RGW-kernel based model outperforms every other neural network, even the CNN model. It is important to note that in this scenario the task has been a lot more difficult than in the previous ones, as the audio signal was subject to heavy background noise. In this case, the interpretability of the RGW-kernel model is especially critical, as it allows us to verify the validity of the trained model.

\begin{table}[!h]
    \centering
    \begin{tabular}{|c|c|c|c|}
        \hline
        \textbf{Model} & \textbf{Mean Acc.} & \textbf{Min Acc.} & \textbf{Max Acc.}  \\ \hline
        WK-NN (proposed) & $\boldsymbol{90.55 \%}$ & $\boldsymbol{87.75 \%}$ & $ \boldsymbol{94.75\%}$ \\ \hline
        RF & $  80.40\%$ & $ 78.00\%$ & $ 82.57\%$ \\ \hline
        SVM-L & $ 62.34\%$& $  61.42\%$ & $ 63.42\%$ \\ \hline
        SVM-RBF & $ 77.71\%$& $  74.85\%$ & $ 80.57\%$ \\ \hline
        NB & $ 49.48\%$& $  48.85\%$ & $ 51.14\%$ \\ \hline
        CNN & $ 89.60\%$& $ 87.00\%$   & $91.25\%$ \\ \hline
        FCNN & $ 55.90\%$& $  53.50\%$ & $ 58.75\%$ \\ \hline
    \end{tabular}
    \caption{Results of detecting drone presence in a noisy environment (scenario 3. in Table~\ref{tab:experiments}).}
    \label{tab:results4}
\end{table}

\section{Conclusion}
\label{sec:conc}
In this study, we introduce a novel model-based neural network, incorporating Rational Gaussian-based wavelet transformation. We have demonstrated the effectiveness of this neural network by solving a series of tasks based around acoustic detection of UAVs in various environments. By using a generic model, we have shown that our scheme can achieve accurate results, and for some tasks, can even surpass the state-of-the art CNN approach, with the use of fewer parameters and small datasets. 

In the future, we plan on using more specialized models in order to achiever higher accuracy on specific tasks, as well as to use these trained models for real time detection and classification.

\section*{Acknowledgment}
This research was funded by the Ministry of Innovation and Technology of Hungary from
the National Research, Development and Innovation Fund, financed under the TKP2021 funding scheme, grant number TKP2021-NVA-03. TD received funding from the Swiss Government Excellence Scholarship No. 2025.0057. This work was supported by the University Excellence Fund of E\"otv\"os Lor\'and University, Budapest, Hungary (ELTE). Project no. K146721 has been implemented with the support provided by the Ministry of Culture and Innovation of Hungary from the National Research, Development and Innovation Fund, financed under the K\_23 "OTKA" funding scheme. This work was supported by the J\'anos Bolyai Research Scholarship of the Hungarian Academy of Science. Supported by the EKÖP-KDP-24 university excellence scholarship program cooperative doctoral program of the Ministry for Culture and Innovation from the source of the National Research, Development and Innovation Fund.

%\begin{appendices}

%\section{Section title of first appendix}\label{secA1}

%An appendix contains supplementary information that is not an essential part of the text itself but which may be helpful in providing a more comprehensive understanding of the research problem or it is information that is too cumbersome to be included in the body of the paper.




%\end{appendices}


\bibliography{sn-bibliography}% common bib file

\section*{Code and data availability}
The python implementation of the proposed methods and experiments can be downloaded from 
\begin{center}
\url{https://gitlab.com/aele02/drone-classification-and-detection/-/tree/main}.
\end{center}
Data that support the findings of this study is available from HUN-REN Centre for Energy research, but restrictions apply to the availability of the data, which was used under license for the current study. The data is not publicly available.
\section*{CRediT Author statement}

\textbf{Gergő Ungvári}: Formal Analysis, Software, Writing - Original Draft. \textbf{Ferenc Braun}: Data Curation, Writing - Original Draft. \textbf{Attila Ámon}: Software, Writing - Review \& Editing. \textbf{Péter Kackst\"adter}: Resources, Writing - Original Draft. \textbf{János Volk}: Supervision, Funding Acquisition,  Writing - Review \& Editing \textbf{Péter Kovács}: Project Administration, Writing - Review \& Editing. \textbf{Tamás Dózsa}: Methodology, Software, Writing - Original Draft.

\section*{Acknowledgment}
This research was funded by the Ministry of Innovation and Technology of Hungary from
the National Research, Development and Innovation Fund, financed under the TKP2021 funding scheme, grant number TKP2021-NVA-03. TD received funding from the Swiss Government Excellence Scholarship No. 2025.0057. This work was supported by the University Excellence Fund of E\"otv\"os Lor\'and University, Budapest, Hungary (ELTE). Project no. K146721 has been implemented with the support provided by the Ministry of Culture and Innovation of Hungary from the National Research, Development and Innovation Fund, financed under the K\_23 "OTKA" funding scheme. This work was supported by the J\'anos Bolyai Research Scholarship of the Hungarian Academy of Science. Supported by the EKÖP-KDP-24 university excellence scholarship program cooperative doctoral program of the Ministry for Culture and Innovation from the source of the National Research, Development and Innovation Fund.
\end{document}